\documentclass[mathematics,article,submit,pdftex,moreauthors]{Definitions/mdpi} 
\firstpage{1} 
\pubvolume{1}
\issuenum{1}
\articlenumber{0}
\pubyear{2022}
\copyrightyear{2022}
\datereceived{} 
\dateaccepted{} 
\datepublished{} 
\hreflink{https://doi.org/} 

\usepackage{fontenc,inputenc,calc,indentfirst,fancyhdr,graphicx,epstopdf, lastpage,ifthen,float,amsmath,setspace,enumitem,mathpazo,booktabs,titlesec,etoolbox,tabto,xcolor,soul,multirow,microtype,tikz,totcount,changepage,attrib,upgreek,amsthm,hyphenat,natbib,footmisc, url,geometry,newfloat,url, geometry, newfloat, caption, pifont}

\makeatletter
\let\c@lofdepth\relax
\let\c@lotdepth\relax
\makeatother
\usepackage{subfigure}
\graphicspath{{./image/}}

\Title{Leveraging Foundation Model Automatic Data Augmentation Strategies and Skeletal Points for Hands Action Recognition in Industrial Assembly Lines}

\TitleCitation{Title}

\Author{Liang~Wu $^{1,2}$, X.-G.~Ma $^{1,2,3,4}$*}

\AuthorNames{Liang~Wu, X.-G.~Ma}

\AuthorCitation{Wu, Liang, X.-G.~Ma}

\address{%
$^{1}$ \quad Faculty of Robot Science and Engineering, Northeastern University, Shenyang, China\\
$^{2}$ \quad Foshan Graduate School, Northeastern University, Foshan, China\\
$^{3}$ \quad College of Information Science and Engineering, Northeastern University, Shenyang, China\\
$^{4}$ \quad The State Key Laboratory of Synthetical Automation for Process Industries, Northeastern University, Shenyang, China\\
}

\corres{Correspondence: maxg@mail.neu.edu.cn}

\abstract{
On modern industrial assembly lines, many intelligent algorithms have been developed to replace or supervise workers. However, we found that there were bottlenecks in both training datasets and real-time performance when deploying algorithms on actual assembly line. Therefore, we developed a promising strategy for expanding industrial datasets, which utilized large models with strong generalization abilities to achieve efficient, high-quality, and large-scale dataset expansion, solving the problem of insufficient and low-quality industrial datasets. We also applied this strategy to video action recognition. We proposed a method of converting hand action recognition problems into hand skeletal trajectory classification problems, which solved the real-time performance problem of industrial algorithms. In the "hand movements during wire insertion" scenarios on the actual assembly line, the accuracy of hand action recognition reached 98.8\%. We conducted detailed experimental analysis to demonstrate the effectiveness and superiority of the method, and deployed the entire process on Midea's actual assembly line.
}
\keyword{Industrial human action recognition, large-scale foundation models, skeleton sequence} 

\preto{\abstractkeywords}{\nolinenumbers} 
\begin{document}
\section{Introduction}
In an era of unprecedented development of intelligent algorithms, existing intelligent industrial assembly lines relied on the algorithms to supervise or replace workers. Action recognition was a core task for these algorithms, and existing research had explored various technical routes for using action recognition algorithms to identify objects or workers from different aspects\;\cite{sturm2023challenges}\cite{sturm2023self}\cite{kellogg2020algorithms}.

Although there had been encouraging results, Industrial human action recognitions (IHARs) were limited in the following aspects. \textbf{datasets}: In the field of IHARs, supervised target detection models, were often required to be trained with a large amount of datasets. The scale, quality, and accuracy of industrial datasets were highly correlated\;\cite{bai2023vision}\cite{agapaki2020cloi}. Some industrial datasets had insufficient data due to difficulties in collection, impact on workers' work, short collection time, and low collection frequency, making it difficult to support the training of advanced algorithms such as deep learning that heavily rely on data. Therefore, achieving low-cost, low-time-consuming, intelligent, and even fully automatic datasets annotation was crucial. \textbf{Real-time performance}: In the field of IHARs, real-time algorithms could detect problems in the assembly process promptly, such as equipment failures and assembly anomalies, and take measures to avoid delays and losses in the assembly process. It could also achieve real-time control of the assembly process, such as the control of automated assembly lines, and improve assembly efficiency and product quality. Therefore, the real-time performance of algorithms was crucial in actual assembly lines, which could help enterprises improve assembly efficiency and product quality and reduce assembly costs\;\cite{ji2021real}\cite{roy2021deep}.

Therefore, in this paper, we proposed a very promising industrial dataset expansion strategy, which could complete high-quality and large-scale industrial datasets for any workpiece or tool with only a minimal amount of manual intervention. We also proposed a hand action recognition approach from video to skeletal points to skeletal point tracking and finally to action classification, and demonstrated its effectiveness and superiority. This innovative approach addressed the challenge of implementing large models without corresponding dataset volume in the industry, as well as achieving high-precision hand action recognition in industrial scenarios. The main contributions of this paper were summarized as follows:

\begin{itemize}
    \item We pioneered a very promising industrial dataset augmentation strategy, using foundation models to achieve efficient, high-quality, and large-scale industrial dataset expansion.
    \item We integrated skeletal point detection into the point tracking method to achieve high-precision tracking of hand skeletal joints. This end-to-end method could obtain hand skeletal point trajectories from videos. Additionally, we incorporated a sliding window mechanism into the time series model to achieve high-accuracy recognition of hand actions from hand skeletal point trajectories.
    \item We deployed the entire process on the actual assembly line of Midea and verified the effectiveness of this method, achieving surprising results.
\end{itemize}
\section{Related Works}
\subsection{Industrial dataset expansion strategies}
A large part of the cost of industrial human action recognition(IHARs) lies in the establishment of the training dataset. High-quality datasets required collection, selection, clipping, expensive annotation, etc., which resulted in high costs for industrial datasets\;\cite{vieira2022low}. Previously, the industry typically used expensive and time-consuming manual annotation methods to complete dataset construction. According to statistics, human annotators could only complete around 150 fine-grained action bounding box annotations in one hour.
While typical deep learning (DL) methods were found to be suitable for achieving low-cost automatic annotations for non-industrial scenarios, such as in\;\cite{cao2020automatic} and\;\cite{alshehri2022deepaia}, there were limited investigations into the use of automatic annotation in complex industrial scenarios.
This issue was initially addressed using large models that have strong generalization abilities, which gradually swept the globe and impacted the entire industry over the past two years. Numerous fusion techniques of LSFMs were proposed for object detection, including GLIP\;\cite{li2022grounded}, and Grounding DINO\;\cite{liu2023grounding}, OWL-ViT\;\cite{minderer2205simple}. The Grounding DINO achieved a 52.5 AP on the small-scale COCO test set, which improved to 63.0 AP after fine-tuning without any training data. Grounded DINO, known for its high recall rate, can be used to filter and detect the target objects in every frame of a video. It can output detection boxes based on simple prompts provided by humans and can identify non-everyday objects, such as electric water heater boxes and welding guns, through human descriptions. Afterward, we only need to manually filter and fine-tune the detection boxes, reducing the workload of annotation. However, there is still a significant gap from a fully automated annotation system, as using only Grounded DINO for detection accuracy is not optimal and still requires a considerable amount of manual filtering to meet the standards of industrial datasets. But our work perfectly solved this problem.

\subsection{Skeleton-based action recognition}
Skeleton-based action recognition has gained increasing attention in recent years due to its ability to capture only the action information while being immune to environmental interference such as background and lighting changes, as it only contains pose information. Within the program, the human skeleton was represented as a list of coordinates\;\cite{duan2022revisiting}. Graph convolutional networks(GCN) were widely applied in skeleton-based action recognition\;\cite{liu2020disentangling}\cite{cheng2020decoupling}. The introduction of GCN into skeleton-based action recognition tasks was first proposed in \;\cite{yan2018spatial}, where ST-GCN was introduced to model spatial and temporal synchronization. Building upon this method,\;\cite{song2020richly} addressed occlusion issues in the task and proposed feature extraction from more active skeletal joints, while\;\cite{li2019spatio}\cite{shi2019two} improved modeling capabilities through self-attention mechanisms.  In addition, many works employed convolutional neural networks for skeleton-based action recognition, but the input for the task cannot leverage the advantages of convolutional networks, rendering these methods less competitive than GCN. Our approach differs from the aforementioned works in that, to better satisfy both real-time requirements and accuracy, we integrated skeleton points into point tracking for action recognition.
\section{Methodology}
In this section, we will discuss several key techniques used in the proposed automatic industrial dataset expansion strategy and Skeleton-based action recognition. First, we will introduce the technologies and overall process used in the industrial dataset expansion strategy. Next, we will present our proposed method, skeleton-based tracking (SBT). Finally, we will discuss the effectiveness of incorporating the sliding window method into action recognition.
\begin{figure}[!t]
    \centering
    \includegraphics[width=0.6\linewidth]{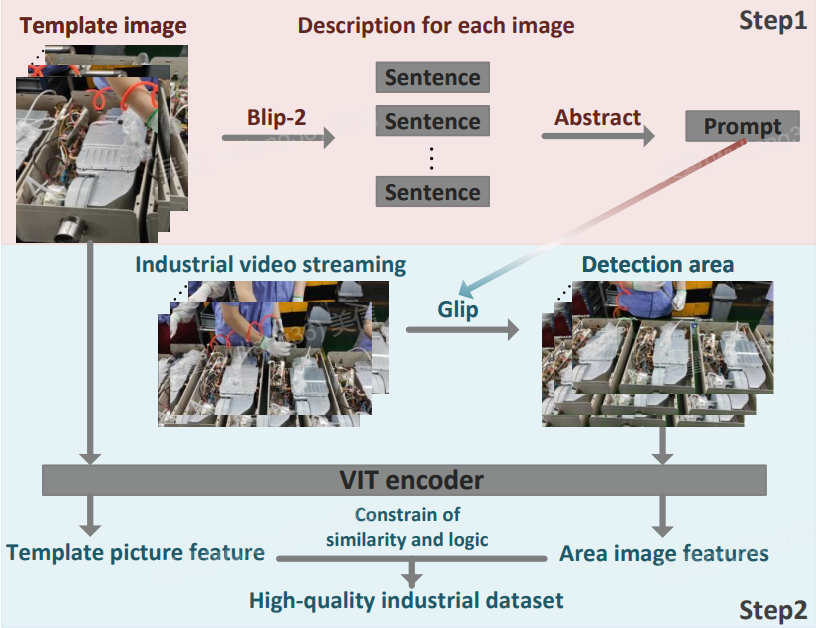}
    \caption{Pipeline of automatic industrial dataset expansion strategy.}
    \label{fig:1}
\end{figure}
\subsection{Automatic industrial dataset expansion strategy}
Our automatic industrial dataset expansion strategy relies on the inference parts of the blip2\;\cite{li2023blip}, glip\;\cite{li2022grounded}, and vit\;\cite{dosovitskiy2020image} models. The role of blip was to identify excellent prompts in the video stream that were beneficial for glip recognition. Glip was responsible for identifying the regions of interest in the images using these prompts. Vit encoder was used to encode the images and filter out the correct augmented dataset.

A simple example was shown in Figure \ref{fig:1}: In the actual assembly line, we needed to detect the \textbf{"water heater metal casing"}. \textbf{Step1:} the collected assembly line video stream was transformed into images, and 10 images of "water heater metal casing" were manually selected as template images. These 10 images were used as input for blip2, which generated a prompt suitable for this scene. Using the prompt selected by blip2 often improved the confidence level of glip recognition. In this scenario, the recognition rate of prompt: \textbf{"machine"} generated by blip2 was higher than the recognition rate of prompt: \textbf{"box"} that people come up with by themselves. This may be because both blip2 and glip encoders are based on the vit structure and have some correlation. \textbf{Step2:} the prompt and all actual assembly line images were input into Glip, which output the target box for each actual assembly line images and its corresponding "water heater metal casing". The actual assembly line images were then cut according to the target boxes. Each cut image and the 10 previously selected template images were input into the vit encoder to obtain the feature maps of the recognition area and the template. The similarity between them was calculated to determine whether these recognition areas were suitable for the training set. Using our automatic expansion strategy, we were able to expand from just 10 manually selected template images to a high-quality industrial dataset consisting of tens of thousands of images. We could then use this industrial dataset to train small models, such as YOLOv5\;\cite{redmon2016you}, specialized for specific scenes.

\subsection{Skeleton-based action recognition}
In addition to using the aforementioned strategy to obtain the dataset for training the specialized detection model, we also needed to train a simple LSTM\;\cite{shi2015convolutional} time-series model for skeleton point action classification. The task pursued real-time and accuracy, and the method used to detect skeleton points was very accurate. Therefore, even with a simple time-series model, the performance achieved in completing the action classification task was surprisingly good. 

\begin{figure}[!t]
    \centering
    \includegraphics[width=0.75\linewidth]{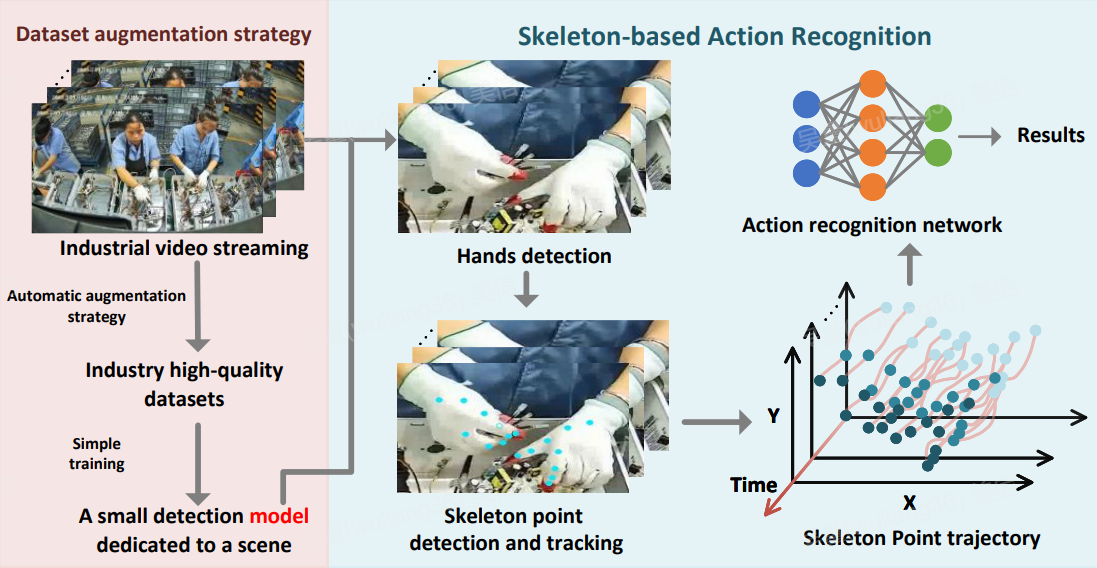}
    \caption{The complete process from data expansion to hand movement recognition of assembly line workers.}
    \label{fig:2}
\end{figure}

An example scenario of workers' \textbf{"hand movements during wire insertion"} on a assembly line is shown in Figure \ref{fig:2}. The red area on the left represented the dataset expansion strategy described in the previous section, which outputted a specialized model for a specific scene. The blue area on the right represented Skeleton-based Action Recognition, which utilized the specialized model on the left to output the action classification ultimately. During the inference process, we proposed the SBT method: first, the real-time assembly line video was input into the specialized detection model to detect the hand box of the worker's starting action. Then, the skeleton points were extracted from this hand box and input into the co-tracker\;\cite{karaev2023cotracker} for tracking, outputting the motion trajectory of the hand skeleton points. Finally, we input these motion trajectories into the trained LSTM to obtain the final action classification result.
\subsection{Sliding window method}
At the end of the model, we incorporated a simple but effective technique called sliding window method. This task was a temporal action recognition task, and considering the requirement of real-time performance, we needed to ensure that the data dimension of the LSTM input was consistent, which could help us to perform batch training and inference. Therefore, we conducted experiments to determine the size of the sliding window and its impact on action recognition accuracy. For example, for an action of inserting and unplugging a cable by a worker, which contained $\alpha=100$ frames, we slid a window of size 16 frames to cover the entire action. We used the LSTM to classify the action based on the $\beta=16$ frames in each window, with a step size of $\gamma=1$. According to Eq.\ref{eq1}, there were $\delta=85$ windows with classification results for this action. We considered the classification result that appeared most frequently among these 85 windows as the final classification result for the entire action. This simple method solved two tricky problems at the same time: first, in actual assembly line actions, the duration of each action varied, which could prevent batch training and inference. By using the sliding window method, we could unify the size of the data to the size of the sliding window. Second, there were many redundant and meaningless noise frames in each complete action, and the sliding window method could reduce the interference of noise and improve the accuracy of action classification.
\begin{equation}\label{eq1}
\frac{\alpha - \beta}{\gamma} + 1= \delta,\beta\in (0,\alpha-\gamma]
\end{equation}
\section{Experiments and Analysis}
\subsection{Dataset preparation}
All of our data was sourced from a representative manufacturing company: Midea. We collected videos with a resolution of 3840*2160 and a size of 8.89GB from the wiring station of its electric water heater, as well as laboratory videos collected at Midea's laboratory. These videos included actions of inserting and unplugging cables. We cut and filtered these original assembly line videos and laboratory videos, and only added manual labels to the laboratory videos as training data.
\subsection{Results of automatic data expansion}
According to the technical route in Figure \ref{fig:2}, we only performed automatic data expansion on the video data from the actual assembly line. We input 8.89GB of actual assembly line videos and selected only 10 images of the moment when the cable insertion action began. Through automatic expansion strategies, we expanded the images to 11,865 in the industrial dataset, of which 97.9\% met our requirements. Subsequently, a small amount of erroneous data was manually selected to achieve a high-quality industrial dataset.
\subsection{Performance evaluation of specialized detection small models}
We trained a large amount of industrial dataset obtained from the previous automatic data expansion method using the mature YOLOv5 model. In our laboratory human hand detection scenario, We compared the performance of the trained specialized small model with the untrained GLIP and Grounding Dino, and the results are shown in Table \ref{table:1}. From the table, we can intuitively see that although the visual large model has strong generalization ability and can achieve an accuracy of about 67\%-69\% without unfine-tuned, it is still far from sufficient for industrial scenarios, and its inference time is long, with only 8 frames per second, which is not conducive to deployment on actual assembly lines. Therefore, we can use these visual large models to expand the dataset and train specialized small models with both accuracy and real-time performance, such as YOLOv5 in the table. Through actual research, we found that the situation on actual assembly lines is not very variable, and the model does not need to have strong generalization ability. In some cases, overfitting to a specific scene is even necessary to improve accuracy. Therefore, our strategy is very effective and promising for actual assembly lines, and we had completed the first step of hand action recognition: detecting the hand.

\begin{table}[h!]
\caption{Detection model performance comparison. \dag\;represents that the results in this column were obtained by testing on an A100.}
\begin{tabular}{cccc}
\hline
Detect Method          & Description  & Speed\;\dag   & Accuracy \\ \hline
Glip                   & unfine-tuned & 6.11FPS & 0.671    \\
Grounding dino         & unfine-tuned & 8.37FPS & 0.694    \\
\textbf{Specialized Model(our)} & \textbf{trained}      & \textbf{42.81FPS} & \textbf{0.893}    \\ \hline
\end{tabular}
\label{table:1}
\end{table}

\subsection{Good properties of skeleton-based tracking(SBT)}
\textbf{Performance}: SBT was able to recognize fine and subtle movements of workers on actual assembly lines at a relatively low cost, as it used a combination of skeleton points and point tracking. In the scenario of recognizing hand gestures, the accuracy of the SBT method could reach 98.8\%. The accuracy of traditional video understanding methods was around 90\%, and required more resources. \textbf{Portability}: SBT only required training a single temporal neural network, so compared to methods that rely solely on image processing for hand gesture recognition, SBT had a shorter training time and was more conducive to portability across different assembly lines and scenarios in the industrial sector. During deployment on actual assembly lines at Midea, we found that SBT reduced porting time and costs by 30\%-35\% compared to traditional industrial action recognition methods. \textbf{Meets industrial scene requirements}: When detecting a specific workpiece or recognizing worker actions on actual industrial assembly lines, a method that has excellent performance, quick portability, and strong stability is required. The SBT method we proposed possessed these characteristics, and we have deployed this method on actual assembly lines with excellent results.
\subsection{Parameter selection of sliding window method}
In the sliding window method, we conducted detailed experiments to determine the parameters in Eq.\ref{eq1}, $\beta$ represents the window size, $\gamma$ represents the step sizes. From Table \ref{table:2}, it can be seen that the accuracy reached the highest point when we selected $\beta=16,\gamma=1$. We conducted experiments with window sizes of 4, 8, and 16 respectively, and found that the larger the window size, the better the model performed. The reason for this is that when the window size is too small, the model may learn less, resulting in unsatisfactory results. We conducted experiments with step sizes of 1, 2, and 4 respectively, and found that the smaller the step size, the better the model performed. The reason for this is that when the step size is too small, the difference between two windows will be reduced, making it easier for the model to learn the temporal information in the video.
\begin{table}[!ht]
\caption{Parameter selection of sliding window method. $\beta$: the window size, $\gamma$: the step sizes.}
\begin{tabular}{c|ccc|ccc|ccc}
\hline
$\beta$           & 4     & 8     & \textbf{16}    & 4     & 8     & 16    & 4     & 8     & 16    \\
$\gamma$           & 1     & 1     & \textbf{1}     & 2     & 2     & 2     & 4     & 4     & 4     \\ \hline
Performance & 0.952 & 0.967 & \textbf{0.988} & 0.949 & 0.954 & 0.976 & 0.942 & 0.644 & 0.969 \\ \hline
\end{tabular}
\label{table:2}
\end{table}

\subsection{Ablation}
In the overall process, there were three places that were relatively innovative and subjective. Firstly, whether to detect the hand first before recognizing hand gestures. From the "\textbf{Cut pictures}" column in Table \ref{table:3}, it can be concluded that adding the step of hand detection can improve the accuracy of gesture recognition by 13.6\%. The reason for this is that detecting the hand can filter out a lot of noise that affects skeletal point detection. Secondly, selecting the number of skeletal points. From the "\textbf{Select skeletal points}" column in Table \ref{table:3}, it can be seen that selecting more active and valid skeletal points is more effective than selecting all skeletal points, with an improvement of 5.1\% in gesture recognition accuracy. The reason for this is that redundant skeletal points not only do not bring benefits, but may also bring noise to the recognition, making the temporal model more difficult to fit. Thirdly, whether to use the sliding window mechanism. From the "\textbf{Sliding window}" column in Table \ref{table:3}, it can be seen that the sliding window mechanism is effective and has a huge improvement of 16.5\%. The reason for this is that the sliding window mechanism can unify videos of different lengths into a fixed length, allowing the temporal model to fit better.
\begin{table}[!h]
\caption{Ablation experiment on SBT method.\textbf{Cut picture}: Perform motion recognition on the detected hand area. \textbf{Select skeletal points}: Select the number of all bone points in the hand to be 40, and the number of bone points in the thumb and index finger (the more active) of the hand to be 18. \textbf{Sliding window}: Use sliding window mechanism.}
\begin{tabular}{ccccc}
\hline
Methods & Cut pictures & Select skeletal points & Sliding window & Performance \\ \hline
SBT     & \ding{56}   & 18                     & \ding{52}     & 0.852       \\
SBT     & \ding{52}   & 40                     & \ding{52}     & 0.937       \\
SBT     & \ding{52}   & 18                     & \ding{56}     & 0.823       \\ \hline
\textbf{SBT}     & \ding{52}   & \textbf{18}                     & \ding{52}     & \textbf{0.988}       \\ \hline
\end{tabular}
\label{table:3}
\end{table}

\section{Conclusion}
We found that constructing training datasets and ensuring the real-time performance of algorithms have long been unresolved challenges in deploying algorithms in the industrial sector. We pioneered a highly promising strategy for augmenting industrial datasets. We integrated skeleton point detection into the point tracking method to achieve high-precision tracking of hand skeletal joints and accomplish the task of recognizing hand movements. Furthermore, we deployed the entire process on Midea's assembly line, achieving satisfactory results.

\authorcontributions{}

\funding{The research was supported by the startup fund of Fashan graduate school, Northeastern University (No. 200076421002), the Natural Science Foundation of Guangdong Province (No. 2020A1515011170), and the 2020 Li Ka Shing Foundation Cross-Disciplinary Research Grant (No. 2020LKSFG05D).}

\dataavailability{The dataset and source code generated during and/or analysed during the current study are available from the corresponding author on reasonable request.}


\conflictsofinterest{No potential conflict of interest was reported by the authors. } 

\begin{adjustwidth}{-\extralength}{0cm}

\bibliography{references}
\bibstyle{mdpi}


%



\end{adjustwidth}
\end{document}